\documentclass[conference]{IEEEtran}
\IEEEoverridecommandlockouts
\usepackage{cite}
\usepackage{amsmath,amssymb,amsfonts}
\usepackage{algorithmic}
\usepackage{graphicx}
\usepackage{textcomp}
\usepackage{xcolor}
\usepackage[T1]{fontenc}
\usepackage{siunitx}
\usepackage[hyphens]{url}
\usepackage{hyperref}
\usepackage{booktabs}
\usepackage{xfrac}
\usepackage{array,multirow,graphicx}
\usepackage{tikz}

\def\BibTeX{{\rm B\kern-.05em{\sc i\kern-.025em b}\kern-.08em
    T\kern-.1667em\lower.7ex\hbox{E}\kern-.125emX}}
\begin{document}

\newcommand{\mycomment}[1]{}
\newcommand\copyrighttext{
	\footnotesize 
	\textcopyright~2023 IEEE. Personal use of this material is permitted. Permission from IEEE must be obtained for all other uses, in any current or future media, including reprinting/republishing this material for advertising or promotional purposes, creating new collective works, for resale or redistribution to servers or lists, or reuse of any copyrighted component of this work in other works. DOI: \href{https://doi.org/10.1109/IV55152.2023.10186653}{10.1109/IV55152.2023.10186653}%
	}%
\newcommand\copyrightnotice{%
    \begin{tikzpicture}[remember picture,overlay]%
 	\node[anchor=south, xshift=0pt, yshift=4pt] at (current page.south)%
 	{\fbox{\parbox{\dimexpr\textwidth-\fboxsep-\fboxrule\relax}{\copyrighttext}}};%
 	\end{tikzpicture}%
}%

\title{
The Impact of Frame-Dropping on Performance and Energy Consumption for Multi-Object Tracking
}

\author{
\IEEEauthorblockN{Matti Henning, Michael Buchholz, and Klaus Dietmayer}
\IEEEauthorblockA{\textit{Institute of Measurement, Control, and Microtechnology} \\
\textit{Ulm University, Germany}\\
{\tt\small \{firstname.lastname\}@uni-ulm.de}}
\thanks{This research is accomplished within the UNICAR\emph{agil} project
(FKZ
16EMO0290). We acknowledge the financial support for the project by the
German Federal Ministry of Education and Research (BMBF).}%
}

\maketitle

\begin{abstract}
The safety of automated vehicles (AVs) relies on the representation of their environment. Consequently, state-of-the-art AVs employ potent sensor systems to achieve the best possible environment representation at all times. 
Although these high-performing systems achieve impressive results, they induce significant requirements for the processing capabilities of an AV's computational hardware components and their energy consumption.

To enable a dynamic adaptation of such perception systems based on the situational perception requirements, we introduce a model-agnostic method for the scalable employment of single-frame object detection models using frame-dropping in tracking-by-detection systems.
We evaluate our approach on the KITTI 3D Tracking Benchmark, showing that significant energy savings can be achieved at acceptable performance degradation, reaching up to 28\% reduction of energy consumption at a performance decline of 6.6\% in HOTA score.
\end{abstract}

\section{Introduction}
\copyrightnotice%
\label{sec:intro}
Environment perception is crucial for the safe operation of automated vehicles (AVs).
To achieve an accurate and robust representation of their environment, state-of-the-art AVs often employ a multi-modal and multi-redundant sensor setup~\cite{Buchholz2021}. 
To model traffic participants, the perception system commonly comprises two central components, following a tracking-by-detection paradigm: 
First, sensor data are processed by high-performance object detection modules.
Then, a tracking algorithm processes the detections to stabilize the object representations over time by using a model for its behavior, e.g., a motion model to track an object's pose. 
In this manner, temporary misdetection of an object due to, e.g., occlusion, can be compensated by model-based predictions. 

The performance of these tracking-by-detection systems is highly dependent on the performance of the employed detection modules. 
This emphasis is reflected in the respective computational requirements, as state-of-the-art detection modules use sophisticated deep-learning-based neural network models. Approaches to reduce the complexity of object detection models have been made, e.g., by adapting their architecture~\cite{liang2021}.
Still, the computational requirements for the tracking component remain significantly lower due to the reduced amount of input data, i.e., object detections instead of raw sensor data. This especially holds for model-based tracking approaches~\cite{Sun2021} compared to deep-learning-based tracking modules~\cite{mabrouk2022, feng2022}.
\begin{figure}[!t]
    \centering
    \includegraphics[width=.9\linewidth]{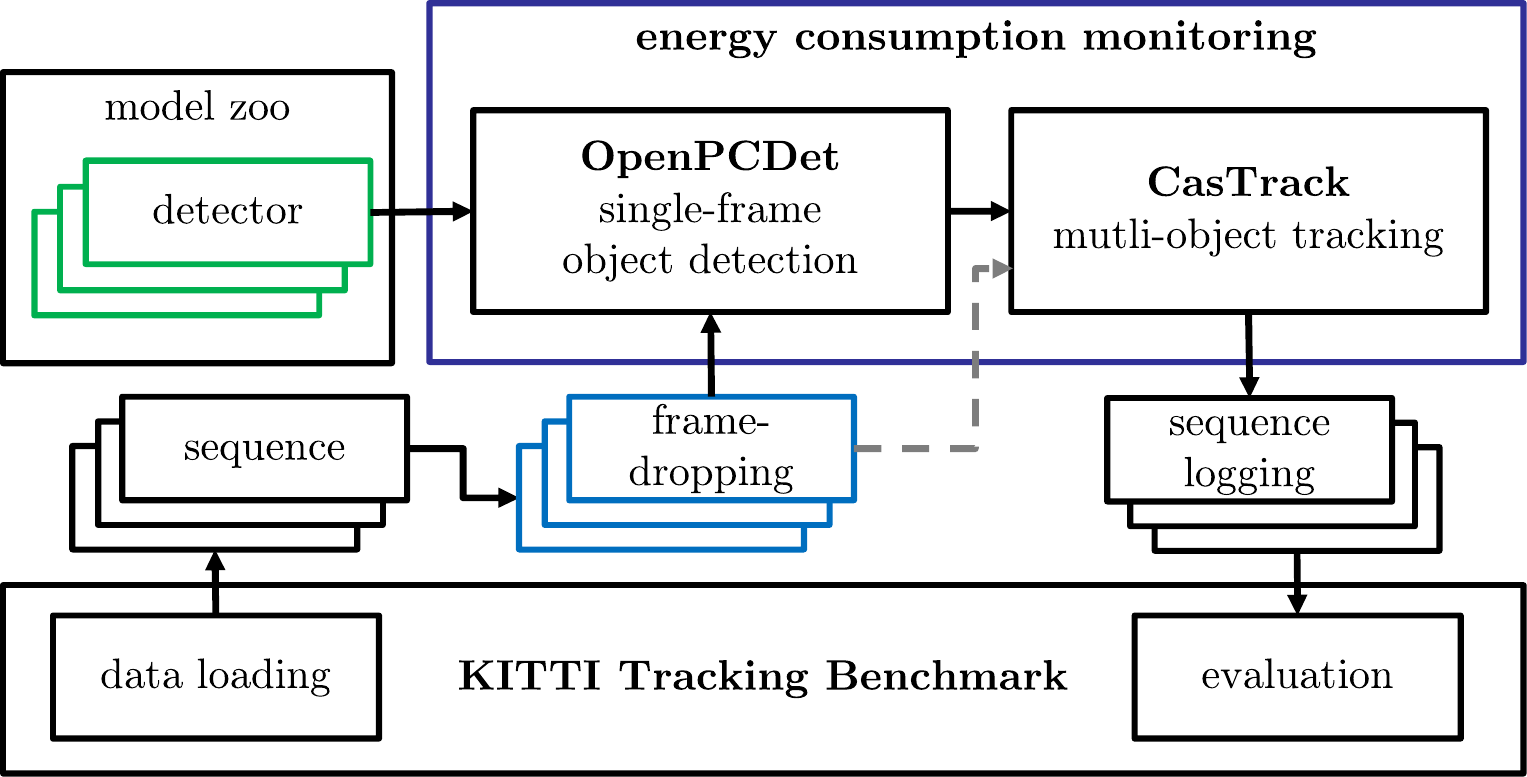}
    \caption{Interaction between dataset, object detection, and tracking. Each detector model and applied frame-dropping is evaluated independently (cf. Section~\ref{sec:method}). The information on currently applied frame-dropping is provided to the tracking module, indicated by the dashed line.}
    \label{fig:intro:overview}
\end{figure}

To enable real-time capabilities of these tracking-by-detection perception systems, state-of-the-art AVs require high-end hardware components of significant energy consumption that decrease the AV's operation time, especially in battery-electric vehicles.
Our recently published concept of situation-aware environment perception (SAEP)~\cite{henning2022, henning2022_UNICAR} aims to reduce the perception system's energy consumption by adapting it based on the situational perception requirements. 
Consequently, the AV's efficiency and operation time can be increased in situations where lower perception performance is acceptable, e.g., in a highway scenario where traffic participants behave more predictably.
However, modern deep-learning-based object detection models are fixed w.r.t. their energy consumption due to their rigid architecture~\cite{henning2022_UNICAR}. Consequently, a dynamic adaptation to the situational requirements often requires changing the employed detection model during runtime, significantly increasing the complexity of the perception system's design.  

In this work, we address this limitation by introducing a method that enables the scalable employment of deep-learning-based single-frame object detection models. 
To do so, we leverage the disparity of computational requirements between object detection and object tracking in tracking-by-detection systems, as well as the model-based prediction capability of the tracking module.
Figure~\ref{fig:intro:overview} presents an overview of our approach, which is further detailed in Section~\ref{sec:method}.
Specifically, we systematically drop sensor data frames for the detection step of the perception system to reduce the perception system's energy consumption. 
The dropped detection frames are compensated by model-based predictions of previously detected traffic participants from the tracking module so that the source frame rate, e.g., of the employed sensor, is restored for the environment representation. 
This approach allows to dynamically adapt to the situational requirements of an AV without any alteration to the employed object detection model.
We focus our work on lidar-based 3D object detectors, as they are especially demanding in terms of computational resources due to the amount and sparsity of their data (compared to camera-based detectors). 
Our main contributions can be summarized as follows:
\begin{itemize}
\item We show that systematic frame-dropping within reasonable margins only slightly decreases the overall perception performance for a tracking-by-detection system.
\item We provide insight into the relationship between frame-dropping and its effect on perception performance and system energy consumption using publicly available perception frameworks.
\item We introduce a model-agnostic method to enable scalable employment of single-frame object detection models 
and derive design recommendations for respective tracking-by-detection perception systems of AVs.
\end{itemize}

%
\section{Related Work}
\label{sec:rw}
Within the context of AVs, frame rates of $\geq$\SI[mode=text]{10}{\hertz} are commonly used for object detection to ensure real-time capability, which is usually defined as processing data in under \SI[mode=text]{100}{\milli\second}.
Our method of frame-dropping (cf. Section~\ref{sec:method} for details) increases the time between consecutively processed data frames and consequently reduces the average detection frame rate.
Although dropping only a few frames has a minor impact on the average detection frame rate, increasing the number of dropped frames eventually relates to tracking in low frame rate (LFR) conditions.
Tracking in LFR conditions commonly focuses on surveillance applications at \SI[mode=text]{5}{\hertz} to \SI[mode=text]{0.5}{\hertz}, where detection modules are often heavily resource-constrained due to  in-field embedded hardware specifications. 
To alleviate the impact of LFR on object detection information, e.g., large positional shifts between frames for the same identities, object similarity features are encoded in the tracking's association step to support continuity~\cite{Wu2021, feng2022}.

A few approaches bridge the gap between real-time and LFR by evaluating the effect of uniform frame rate reduction on the resulting tracking performance using subsampling. 
\cite{mabrouk2022} evaluates subsampled frame rates ranging from \SI[mode=text]{25}{\hertz} to \SI[mode=text]{1}{\hertz} for deep-learning-based tracking methods using camera-based object detection models. The authors show a significant performance drop below \SI[mode=text]{10}{\hertz}, concluding that the evaluated tracking approaches are not suited for lower frame rates. 
Their results are backed by \cite{feng2022}, which similarly shows that current deep-learning-based tracking approaches are incapable of adequately dealing with frame rates $\leq$\SI[mode=text]{2}{\hertz} due to training-induced dependencies on the expected, i.e., trained, frame rate. The authors present an extension of state-of-the-art deep-learning tracking approaches w.r.t. frame-rate agnostic training, as well as the introduction of object similarity features in their association, significantly outperforming existing approaches.

Both of these works primarily refer to the context of pedestrian surveillance using cameras, where significant occlusion and less-predictable motion are expected. 
Further, neither provides insight into the connection of frame rate to computational requirements.
Within the scope of our work, we adapt the presented uniform subsampling approach to frame-dropping (cf. Section~\ref{sec:method}) for the context of AV perception. 
This is the first approach to connect systematic frame-dropping, perception performance, and energy consumption analysis for the context of AV object detection.

%
\section{Method}
\label{sec:method}
The core drive of our work is to gain insight into the relationship between perception performance and system energy consumption dependent on systematically dropping data frames in the detection step of multi-object tracking for AVs. 
It is expected to observe a reduced perception performance at an increased number of dropped frames (cf. Section~\ref{sec:rw}). Consequently, we evaluate the impact of frame-dropping relative to its baseline, i.e., processing every frame of the dataset.
The elements of our method, as presented in Fig.~\ref{fig:intro:overview}, are further detailed in this section.  

\subsection{Employed Frameworks}
\label{sec:frameworks}
AV perception of traffic participants is thoroughly researched, while continued progress is made in performance. Several publicly available datasets, including ground-truth label data of various sensor modalities, are available for training and evaluation, e.g., KITTI~\cite{Geiger2012CVPR}, or nuScenes\cite{nuscenes2019}, which focus on single-frame object detection. 
Besides, specific datasets and evaluation frameworks also exist for the tracking component, both outside the context of automated vehicles, e.g., MOT-Challenge~\cite{MOT16, MOTChallenge20}, as well as within, e.g., KITTI Tracking Benchmark~\cite{Geiger2012CVPR}.
Each of these benchmarks provides an extensive ranking of evaluated methods.

For our work, we apply the lidar-based detection framework OpenPCDet~\cite{openpcdet2020} in combination with the 3D multi-object tracking framework CasTrack~\cite{CasTrack} to the KITTI Tracking Benchmark dataset\cite{Geiger2012CVPR}. 
OpenPCDet integrates various state-of-the-art object detectors, enabling an efficient comparison between them. CasTrack leverages a simplified approach from~\cite{Wu2021} and is especially suited for misdetections. Further, CasTrack provides tight integration of the KITTI Tracking Benchmark evaluation methods via~\cite{luiten2020trackeval}. 
We deliberately refrain from a deep-learning-based tracking approach as per the indicated shortcomings (cf. Section~\ref{sec:rw}). 
By choosing publicly available perception frameworks, we aim to reduce our work's application and reproduction efforts. 

\subsection{Object Detection and Tracking}
\label{sec:method:dettrack}
To evaluate the performance vs. energy consumption trade-off in multi-object tracking, we process the described dataset as per the indicated frame-dropping from Table~\ref{tab:method:subsampling} (cf. Section~\ref{sec:method:data}) with a selection of available object detection models using OpenPCDet~\cite{openpcdet2020}. The chosen models are PV-RCNN~\cite{shi2020_pvrcnn}, Point-RCNN~\cite{shi2019pointrcnn}, SECOND~\cite{yan2018second}, and PointPillars~\cite{lang2019pointpillars}. The model selection represents well-performing single-frame detector models of different complexity using voxel-based, point-based, and hybrid backbones.

Their detections are forwarded to the CasTrack~\cite{CasTrack} module, which we adapted to generate model-based object predictions in frames where the detection step is dropped.
Consequently, an output for tracked objects is generated in every frame instead of only for frames processed by the detection model, allowing for a consistent evaluation of performance degradation for the applied frame-dropping as per Table~\ref{tab:method:subsampling}. 

In addition to the object detection models, the ground-truth label data is directly provided to the tracking module, representing a perfect detector model at processed frames. This model, labeled GT, generates a reference for the effect of performance reduction due to frame-dropping in multi-object tracking. 
As the object detections are loaded from the dataset, and no actual object detection model is employed for data processing, monitoring the system energy consumption is neglected for the GT detection model.

The combination of detection models and applied frame-dropping results in $(\text{4}+\text{1})\times\text{6}=\text{30}$ variants for evaluation.

\subsection{Data Handling}
\label{sec:method:data}
The KITTI Tracking Benchmark consists of 21 training sequences and 29 test sequences. Since ground truth labels are not available for the KITTI test sequences, we evaluate the performance on the validation split of the training dataset, containing 11 sequences. 

To evaluate the effects of frame-dropping, we adapt the detection step of the perception system according to Table~\ref{tab:method:subsampling} so that only $n$ out of $m$ consecutive lidar frames are processed. 
This approach can be easily applied to online data processing of AVs.
The applied frame-dropping reflects a targeted percentage of processed data. However, as the length of the sequences might not be an exact multiple of $m$, the target percentage is not matched precisely (e.g., \SI[mode=text]{90.25}{\percent} instead of \SI[mode=text]{90}{\percent}). This deviation is consistent for all evaluated variants and will be neglected in Section~\ref{sec:eval}.
For the remainder of this work, the applied frame-dropping is referred to by its processing target.
\begin{table}[t]
    \centering
    \caption{Applied frame-dropping to process $n$ out of $m$ frames.}
    \begin{tabular}{rcccccc}
    \toprule
    processing target & \SI[mode=text]{100}{\percent} &\SI[mode=text]{90}{\percent} & \SI[mode=text]{75}{\percent} & \SI[mode=text]{50}{\percent} & \SI[mode=text]{25}{\percent} & \SI[mode=text]{10}{\percent}\\
    \sfrac{$n$}{$m$} & \sfrac{1}{1} & \sfrac{9}{10} & \sfrac{3}{4} & \sfrac{1}{2} &  \sfrac{1}{4} &  \sfrac{1}{10}\\
    processed frames & 3898         & 3500          & 2919         & 1950         & 979           & 393\\
    \bottomrule
    \end{tabular}
    \label{tab:method:subsampling}
\end{table}

After data loading, the sequences and their respective frames are fed sequentially to the perception system (cf. Section~\ref{sec:method:dettrack}). 
In parallel, the energy consumption of the perception system is monitored, and respective data is stored alongside resulting object tracking results for each sequence for post-processing evaluation (cf. Section~\ref{sec:method:eval}).
For a fair comparison of the monitored system energy consumption of the evaluated variants, we emulate an online-like behavior of data processing  by introducing a cycle time of $t=\SI[mode=text]{100}{\milli\second}$ between frames. In this manner, dropped frames will reflect consistently in the monitored system energy consumption. 
However, larger detection models like Point-RCNN and PV-RCNN require larger inference times. 
Where less complex detection models would wait at an idle state for the next frame to be provided for processing, it is expected that these complex models extend the \SI[mode=text]{100}{\milli\second} cycle time so that the processing of the next frame is delayed. 
Consequently, the reduction in system energy consumption for these complex models at lower processing targets is expected to be slightly lower. 
Still, as we aim for real-world similarity, we adhere to this common assumption for real-time capability.
For online applications requiring real-time capabilities of complex models, model adaptations trading inference speed vs. accuracy might be adopted~\cite{liang2021}.

\begin{table*}[t]
    \centering
    \caption{Evaluation results for the selected detection model variants. Processing target in \si{\percent}. MOTA in \si{\percent}, MOTP in \si{\percent}, median system draw in \si{\watt}, yield in \si{\watt} per reduced point in HOTA score. Highest achieved yield \textbf{bold}, second-highest yield \underline{underlined}.}
    \label{tab:eval:res}
    \setlength\tabcolsep{.1cm}
\begin{tabular}{rl|cccccc|cccccc|cccccc|cccccc}
\toprule
&&\multicolumn{6}{c|}{Point-RCNN}&\multicolumn{6}{c|}{PV-RCNN}&\multicolumn{6}{c|}{SECOND}&\multicolumn{6}{c}{PointPillars}\\
&target& 100& 90& 75& 50& 25& 10& 100& 90& 75& 50& 25& 10& 100& 90& 75& 50& 25& 10& 100& 90& 75& 50& 25& 10\\
\midrule
$\uparrow$&MOTA& 79.4& 78.3& 76.1& 74.5& 52.7& 31.2& 83.7& 82.5& 80.4& 77.8& 54.8& 32.5& 83.2& 81.1& 79.0& 76.7& 52.8& 29.0& 81.2& 79.7& 77.8& 76.2& 53.4& 30.9\\
$\uparrow$&MOTP& 87.1& 87.0& 86.5& 85.9& 84.1& 81.2& 88.5& 88.4& 88.0& 87.6& 85.8& 82.3& 87.4& 87.2& 86.9& 86.5& 84.7& 81.0& 87.8& 87.5& 87.1& 86.5& 84.6& 81.4\\
$\uparrow$&HOTA& 72.3& 71.0& 69.5& 66.8& 56.5& 42.7& 77.9& 77.3& 75.5& 72.6& 62.5& 46.1& 77.1& 75.7& 73.6& 72.0& 60.9& 44.5& 74.9& 74.0& 72.5& 70.2& 59.4& 42.8\\
$\downarrow$&Sys. draw& 304& 290& 271& 240& 204& 180& 314& 306& 286& 253& 210& 178& 270& 254& 228& 194& 172& 156& 213& 208& 199& 184& 164& 155\\
\midrule
$\uparrow$&yield& -& 11.3& 12.1& 11.7& 6.4& 4.2& -& \underline{14.6}& 12.1& 11.7& 6.8& 4.3& -& 11.7& 12.1& \textbf{15.0}& 6.1& 3.5& -& 6.5& 6.2& 6.2& 3.2& 1.8\\
    \bottomrule
    \end{tabular}
  \end{table*}

\begin{table}
    \centering
    \caption{Evaluation results for the GT variants. System draw measurements are neglected (cf. Section~\ref{sec:method:dettrack}).}
    \label{tab:eval:resGT}
    \setlength\tabcolsep{.1cm}
    \begin{tabular}{rl cccccc}
    \toprule
    &&\multicolumn{6}{c}{GT}\\
    &target in \si{\percent}& 100& 90& 75& 50& 25& 10\\
    \midrule
    $\uparrow$&MOTA in \si{\percent}& 98.8& 97.6& 95.8& 93.4& 68.4& 44.3\\
    $\uparrow$&MOTP in \si{\percent}& 97.2& 96.7& 96.1& 94.9& 91.5& 87.0\\
    $\uparrow$&HOTA                 & 98.0& 96.1& 93.6& 90.3& 72.8& 56.7\\
    \bottomrule
    \end{tabular}
  \end{table}
  
\subsection{Evaluation}
\label{sec:method:eval}
For the evaluation of the perception performance, we leverage the evaluation framework from CasTrack~\cite{CasTrack, luiten2020trackeval}, adhering to the KITTI Tracking Benchmark evaluation metrics of HOTA~\cite{Luiten2020} and CLEAR~\cite{bernardin2008}. For the conciseness of our evaluation, we restrict the presented results to the car class. 

To evaluate the effect of frame-dropping on the computational requirements, we measure the overall system energy consumption, i.e., the system power draw, during the emulated online-like processing using an external measurement device, averaging from 100 samples per second. The reported results refer to the median of the averaged results.

To provide means for a comparison between evaluated detection models (cf. Section~\ref{sec:method:dettrack}), we introduce the \textit{yield} of a model, 
\begin{align}
    \text{yield}^\text{target}_\text{model} &= \frac{\text{system draw}_\text{model}^\text{100}-\text{system draw}_\text{model}^\text{target}}{\text{HOTA}_\text{model}^\text{100}-\text{HOTA}_\text{model}^\text{target}}\,,
\end{align}
representing the reduction of system draw in \si{\watt} per reduced point in HOTA score, each in relation to their respective baseline, i.e., processing target of \SI[mode=text]{100}{\percent}. Except for the system draw, where smaller values are better, all other metrics show better behavior at larger values.

%
\section{Evaluation}
\label{sec:eval}
The dataset is processed in Python on a consumer-grade PC running Ubuntu. Energy consumption monitoring refers to this system equipped with an AMD Ryzen Threadripper 2990WX CPU and an Nvidia RTX 2080Ti 11GB GPU on 64GB RAM.

The results for the chosen detector model variants are provided in Table~\ref{tab:eval:res}, supplemented by the GT results in Table~\ref{tab:eval:resGT}.

\subsection{Performance of the GT Detector}
\label{sec:eval:quant}
The reference results in Table~\ref{tab:eval:resGT} verify the results of~\cite{mabrouk2022} and~\cite{feng2022}, showing a decline in HOTA score with an increasing percentage of dropped frames. While the decline for \SI[mode=text]{25}{\percent} and \SI[mode=text]{10}{\percent} processing target is significant, processing targets of \SI[mode=text]{90}{\percent} or \SI[mode=text]{75}{\percent} only result in a minor HOTA score decline. 
Relating the reported HOTA scores to the corresponding MOTA and MOTP values of their respective processing targets, it is apparent that the score reduction is referring more to a decrease in accuracy, i.e., misdetections/-associations or identity switches, and less to a decrease in precision, i.e., similarity of object detections to the ground-truth object labels. 
This behavior is consistent with our expectation for lowering the processing target: Known  object detections will be stabilized according to their model-based predictions, resulting in a high precision score. Missing objects will remain missing for continuously dropped frames until the first processed frame provides a detection so that accuracy will drop further with lower processing targets, i.e., increasing numbers of dropped frames.
Consequently, the presented results verify the functionality of the employed tracking adaptations, achieving near-perfect scores at a processing target of \SI[mode=text]{100}{\percent}. The divergence from a perfect score relates to expected tracking effects, e.g., due to object occlusions, detection to track mismatches, or object predictions extending the span of object labels.

\subsection{Performance of the Object Detection Models}
The performance results in Table\ref{tab:eval:res} are achieved after adapting the parameters for both object detection and tracking, starting from their default parameterization from OpenPCDet~\cite{openpcdet2020}, and CasTrack~\cite{CasTrack}, respectively. 
For the context of our work, this is especially important for object creation and deletion thresholds at the respective processing targets.
The achieved HOTA scores for PV-RCNN and SECOND at \SI[mode=text]{100}{\percent} target are similar, although slightly lower than the reported scores on~\cite{CasTrack} (\SI[mode=text]{-0.4}{} for PV-RCNN, \SI[mode=text]{-0.2}{} for SECOND). The achieved HOTA score for Point-RCNN at \SI[mode=text]{100}{\percent} is \SI[mode=text]{4.3}{} points lower. We conclude that our system achieves comparable results, as the provided model checkpoints in OpenPCDet differ from the ones used in CasTrack.
Following the conclusions from~\cite{feng2022} w.r.t. frame-rate dependency in tracking applications as well as the implications of~\cite{ravikiran2020} w.r.t. parameter optimization for low-frame-rate tracking applications, we expect further potential in perception performance. 
However, focusing on the comparison between the evaluated variants, further optimization is outside of the scope of this work.

\begin{figure}[t]
    \centering
    \includegraphics[width=1\linewidth]{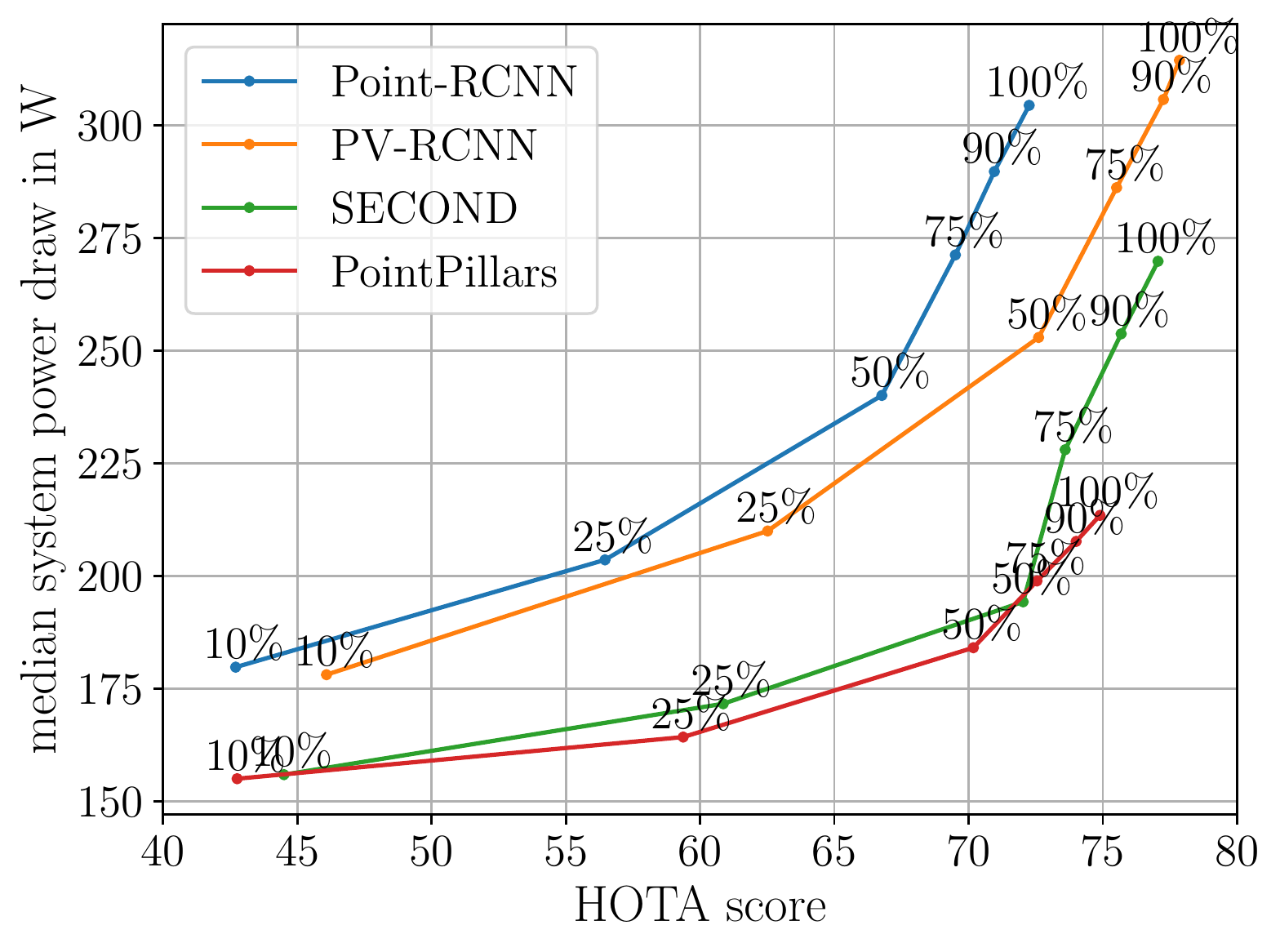}
    \caption{Achieved HOTA score vs. induced system draw for all detection model variants from Table~\ref{tab:eval:res}. Percentages refer to processing targets.}
    \label{fig:eval:hota_draw}
\end{figure}

The results from Table~\ref{tab:eval:res} are visualized in Fig.~\ref{fig:eval:hota_draw}.
The more complex detection models Point-RCNN and PV-RCNN induce a significantly higher system draw than PointPillars and SECOND. 
Further, Point-RCNN achieves lower HOTA scores for inducing considerably higher system draw, whereas PV-RCNN achieves the highest HOTA score at \SI[mode=text]{100}{\percent} at a similar system draw. SECOND follows quickly in terms of achieved HOTA score at a considerably lower system draw. 
PointPillars achieves lower HOTA scores at a similarly lower system draw.
Reducing the processing target consistently lowers both the HOTA score as well as the system draw for all evaluated detection models. 
These results indicate that systematic and random errors in our approach to system draw monitoring due to unrelated processes running on the system are neglectable.
Relating the reported HOTA scores to the corresponding MOTA and MOTP values in Table~\ref{tab:eval:res} shows identical behavior to the GT baseline, where MOTP remains consistently higher than MOTA over reduced processing targets.

\begin{figure}[t]
    \centering
    \includegraphics[width=1\linewidth]{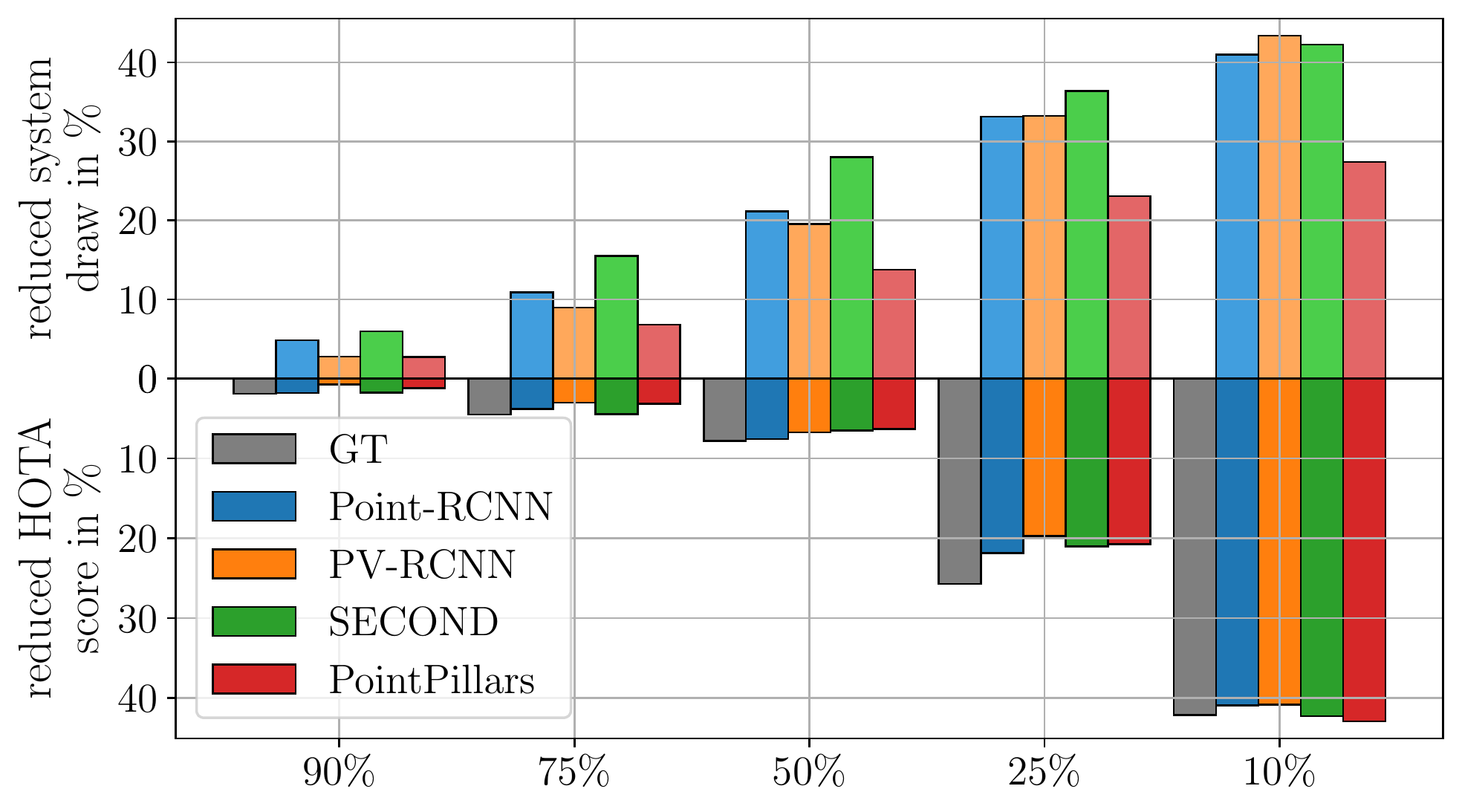}
    \caption{Relative reduction in HOTA score and system draw at the indicated processing targets w.r.t. the baseline value at \SI[mode=text]{100}{\percent} processing target.}
    \label{fig:eval:bar_comp}
\end{figure}
Figure~\ref{fig:eval:bar_comp} presents the relative reduction in achieved HOTA score as well as in reduced system draw compared to the respective baseline processing target of \SI[mode=text]{100}{\percent}.
First, the relative reduction in HOTA score for the evaluated object detection models behaves similarly to the corresponding relative reduction of the GT baseline, as well as compared between each other. Again, the lower the processing target, the larger the reduction in the HOTA score. 
In all cases, the GT decreases slightly more, reflecting the similarly higher decrease in absolute MOTP score (cf. Table~\ref{tab:eval:res} and~\ref{tab:eval:resGT}).  
Further, the more complex detection models generally lead to a higher reduction in system draw with lower processing targets. 
Although this general behavior still holds for PointPillars, which induces the lowest system draw, the relative reduction in system draw decreases with lower processing targets. 
This effectively represents the lower bound of the system draw at its idle state.

\begin{figure}[t]
    \centering
    \includegraphics[width=1\linewidth]{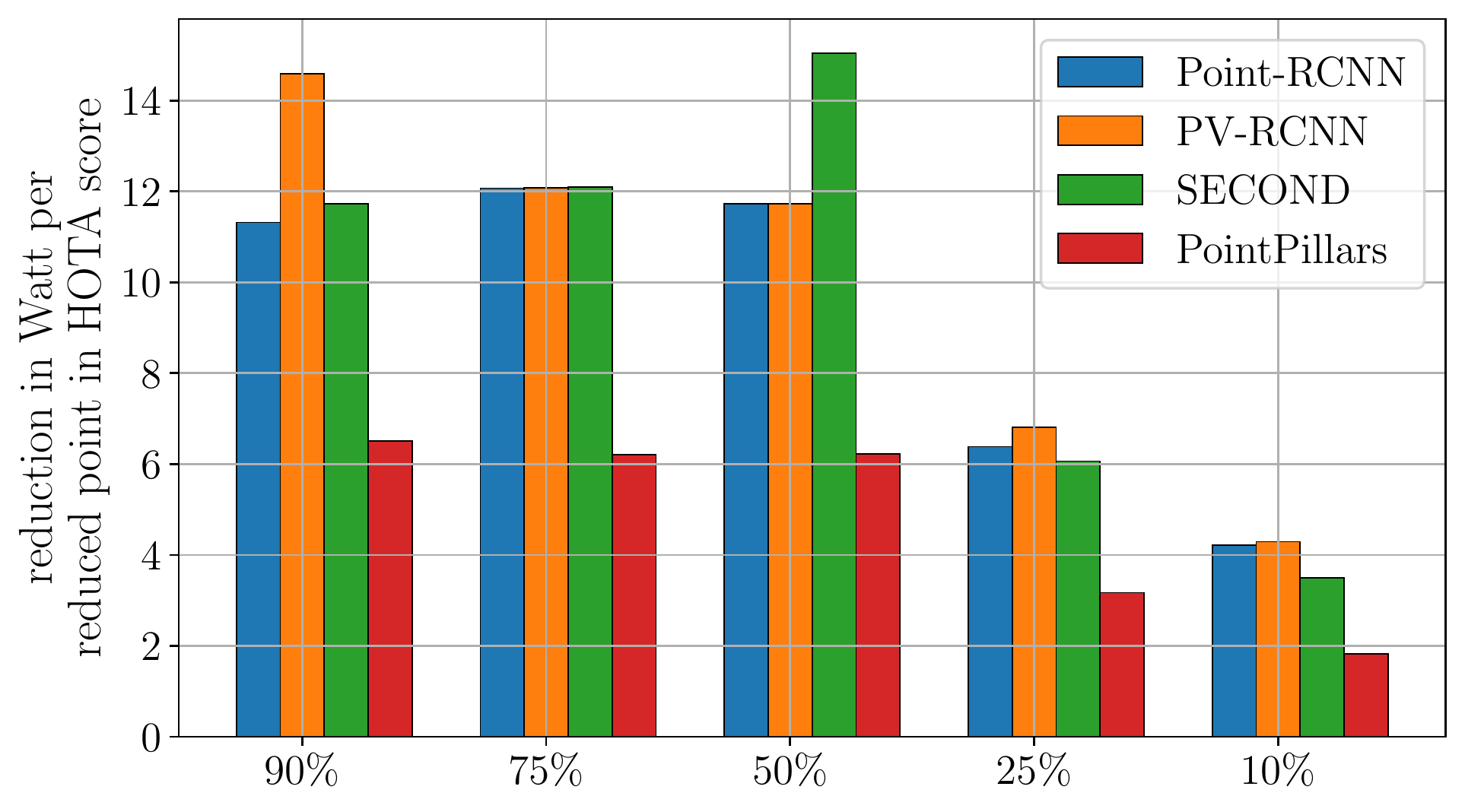}
    \caption{Yield of the evaluated object detection models at the indicated processing targets.}
    \label{fig:eval:bar_yield}
\end{figure}

Lastly, Fig.~\ref{fig:eval:bar_yield} presents the resulting yield of the evaluated object detection models, combining the respective relative decrease in HOTA score and system draw w.r.t. the baseline processing target at \SI[mode=text]{100}{\percent}.
The highest yield is achieved by SECOND at \SI[mode=text]{50}{\percent} (\SI[mode=text]{15.0}{}), referring to a reduction in system draw of \SI[mode=text]{76}{\watt} at a decline of approximately~\SI[mode=text]{5.1}{} points in HOTA score (\SI[mode=text]{28.1}{\percent} for \SI[mode=text]{6.6}{\percent}). 
PV-RCNN follows this yield closely at \SI[mode=text]{90}{\percent} (\SI{14.9}{}), referring to a reduction in system draw of \SI[mode=text]{8}{\watt} at a decline of approximately~\SI[mode=text]{0.6}{} points in HOTA score (\SI[mode=text]{2.5}{\percent} for \SI[mode=text]{0.8}{\percent}).
PointPillars consistently provides the lowest yield, reflecting its generally low system draw approaching the system's idle state.
Interestingly, the yield for most of the evaluated detection models is very similar, except for PointPillars as an exceptionally efficient detector model with low GPU load. 
Further, the yield decreases sharply for processing targets below \SI[mode=text]{50}{\percent}, representing the increasing corresponding performance drops. 
Consequently, reducing the processing target within reasonable boundaries, e.g., up to  \SI[mode=text]{50}{\percent}, provides a significant reduction in energy consumption, i.e., system power draw, at a reasonable performance cost.

\subsection{Design Recommendations for a Scalable Perception System}
\label{sec:eval:saep}
Considering the presented results, recommendations can be derived for designing a perception system that dynamically adapts according to its situational requirements. 
As an example, for a system aiming for low energy consumption, PointPillars presents a reasonably well-performing model at a low system draw. Although its yield is the lowest, further reduction of energy consumption is possible. 
Another system aiming for the possibility of high-performing perception might employ a model similar to PV-RCNN at a default data processing target of \SI[mode=text]{50}{\percent}.
While the system draw is generally higher, it enables the system to scale towards higher performance in critical situations at the cost of increased system draw.
Further, reducing the processing target below \SI[mode=text]{50}{\percent} might be accompanied by continued adaptation and optimization of the tracking module regarding methods for tracking in low frame rate conditions (cf. Section~\ref{sec:rw}).

Our recently published concept of SAEP~\cite{henning2022} provides means for the dynamic assessment of the situation of an AV to identify the corresponding requirements to the perception system, as well as means to adapt the perception system accordingly. 
In this context, this work presents a model-agnostic method for employing single-frame object detectors in a scalable manner, significantly reducing the design complexity of adaptable AV perception systems. 
Further, the proposed method requires no model adaptations.

\subsection{Mitigating Late Object Detection}
\begin{figure}[t]
    \centering
    \includegraphics[width=1\linewidth]{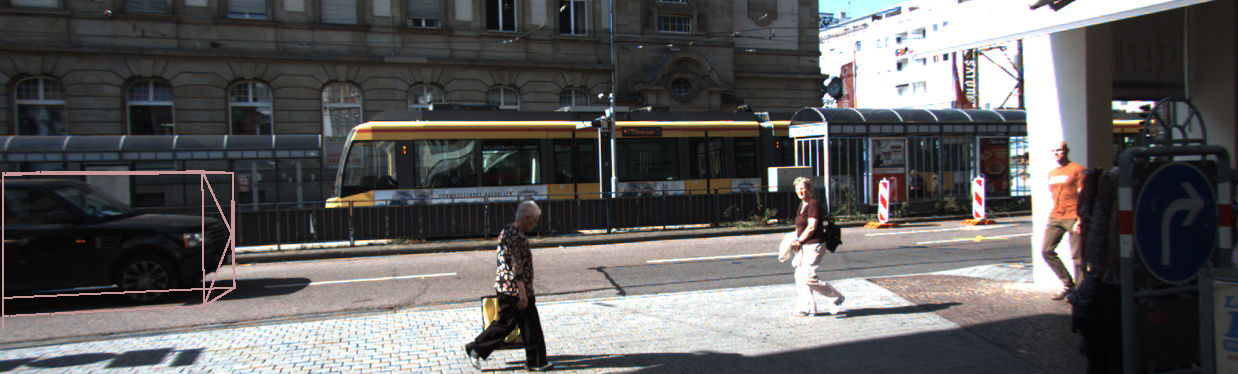}
    \caption{Example for a potentially risk-inducing situation due to late object detection at \SI[mode=text]{10}{\percent} processing target. Object labels for cars are indicated in red.}
    \label{fig:eval:qual_late}
\end{figure}

Although the presented approach provides the potential for a significant reduction in energy consumption of perception systems at a reasonable performance loss, it induces potential risks regarding the safety of AVs that are not directly represented in an averaged performance metric such as HOTA. 
Specifically, by dropping consecutive frames from the data source's base frame rate, the detection frame rate effectively decreases. Consequently, detecting an object that requires an immediate reaction might be delayed, hampering a timely system reaction. 
An example of a potentially risk-inducing missing detection (object label indicated in red) appearing behind an occluding building on the left-hand side (not visible in the camera view) is presented in Fig.~\ref{fig:eval:qual_late}. 
To mitigate the induced risk of delayed object detection, a lower bound on the accepted data processing target might be employed, limiting both the worst-case delay, as well as the potential of our presented method.

An alternative approach is to evaluate the potential threat of an AV's current situation on a lower perception level. 
An identified threat can then trigger the immediate processing of the next available frame, effectively reducing the induced delay due to frame-dropping while maintaining the full potential for reduced energy consumption in non-critical situations. 
Low-level threat region identification methods are presented by\cite{henning2022_threat, Zhou2022}, which neatly fit into our concept of SAEP~\cite{henning2022}.

\section{Conclusions}
In this work, we introduced a model-agnostic method for the scalable employment of single-frame deep-learning-based lidar object detection models that requires no adaptations to the employed models. To do so, we leveraged the tracking module's model-based object predictions to compensate for dropped frames in the detection step of a tracking-by-detection perception system.

Employing off-the-shelf open-source frameworks for object detection and tracking, we assessed the impact of frame-dropping on the system power draw and perception performance. We presented an extensive evaluation of our method on the KITTI Tracking Benchmark dataset. 
The evaluated object detection models yield up to \SI[mode=text]{15.0}{}\si{\watt} per reduced point in HOTA score, showing that a significant reduction in energy consumption can be achieved with a reasonable decline in perception performance. 
To efficiently leverage our method's provided scalability, situation-aware environment perception can identify a reasonable decline and adapt the AV's perception system accordingly.

In future work, we aim to test our method on real AVs during runtime while extending our selection of object detection models and tracking approaches.
Besides, we have briefly outlined the potential risk of our method regarding safety-critical situations due to late object detection. To gain further insight into the corresponding real-world implications, we aim to investigate the effect of non-constant frame-dropping, reflecting an external trigger to process the next possible frame. 

\bibliographystyle{IEEEtran} 
\bibliography{IEEEabrv,root}

\begin{thebibliography}{10}
\providecommand{\url}[1]{#1}
\csname url@samestyle\endcsname
\providecommand{\newblock}{\relax}
\providecommand{\bibinfo}[2]{#2}
\providecommand{\BIBentrySTDinterwordspacing}{\spaceskip=0pt\relax}
\providecommand{\BIBentryALTinterwordstretchfactor}{4}
\providecommand{\BIBentryALTinterwordspacing}{\spaceskip=\fontdimen2\font plus
\BIBentryALTinterwordstretchfactor\fontdimen3\font minus
  \fontdimen4\font\relax}
\providecommand{\BIBforeignlanguage}[2]{{%
\expandafter\ifx\csname l@#1\endcsname\relax
\typeout{** WARNING: IEEEtran.bst: No hyphenation pattern has been}%
\typeout{** loaded for the language `#1'. Using the pattern for}%
\typeout{** the default language instead.}%
\else
\language=\csname l@#1\endcsname
\fi
#2}}
\providecommand{\BIBdecl}{\relax}
\BIBdecl

\bibitem{Buchholz2021}
M.~Buchholz, J.~Müller, M.~Herrmann, J.~Strohbeck, B.~Völz, M.~Maier,
  J.~Paczia, O.~Stein, H.~Rehborn, and R.-W. Henn, ``{Handling Occlusions in
  Automated Driving Using a Multiaccess Edge Computing Server-Based Environment
  Model From Infrastructure Sensors},'' \emph{IEEE Intell. Transp. Syst. Mag.},
  vol.~14, no.~3, pp. 106--120, 2022.

\bibitem{liang2021}
T.~Liang, J.~Glossner, L.~Wang, S.~Shi, and X.~Zhang, ``{Pruning and
  Quantization for Deep Neural Network Acceleration: A Survey},''
  \emph{Neurocomputing}, vol. 461, pp. 370--403, 2021.

\bibitem{Sun2021}
Z.~Sun, J.~Chen, L.~Chao, W.~Ruan, and M.~Mukherjee, ``{A Survey of Multiple
  Pedestrian Tracking Based on Tracking-by-Detection Framework},'' \emph{IEEE
  Trans. on Circuits and Syst. for Video Technol.}, vol.~31, no.~5, pp.
  1819--1833, 2021.

\bibitem{mabrouk2022}
A.~Y.~B. Mabrouk, G.~Facciolo, R.~G. von Gioi, and A.~Davy, ``{An assessment of
  Multi Object Tracking on low framerate conditions},'' \emph{hal-03641298},
  2022.

\bibitem{feng2022}
W.~Feng, L.~Bai, Y.~Yao, F.~Yu, and W.~Ouyang, ``{Towards Frame Rate Agnostic
  Multi-Object Tracking},'' \emph{arXiv:2209.11404}, 2022.

\bibitem{henning2022}
M.~Henning, J.~Müller, F.~Gies, M.~Buchholz, and K.~Dietmayer,
  ``{Situation-Aware Environment Perception Using a Multi-Layer Attention
  Map},'' \emph{IEEE Trans. on Intell. Veh.}, vol.~8, no.~1, pp. 481--491,
  2023.

\bibitem{henning2022_UNICAR}
M.~Henning, M.~Buchholz, and K.~Dietmayer, ``{Situation-Aware Environment
  Perception for Decentralized Automation Architectures},'' in \emph{IEEE
  Intell. Veh. Symp.}, 2022, pp. 1087--1092.

\bibitem{Wu2021}
H.~Wu, W.~Han, C.~Wen, X.~Li, and C.~Wang, ``{3D Multi-Object Tracking in Point
  Clouds Based on Prediction Confidence-Guided Data Association},'' \emph{IEEE
  Trans. on Intell. Transp. Syst.}, 2021.

\bibitem{Geiger2012CVPR}
A.~Geiger, P.~Lenz, and R.~Urtasun, ``{Are we ready for Autonomous Driving? The
  KITTI Vision Benchmark Suite},'' in \emph{Conf. on Computer Vision and
  Pattern Recognition}, 2012.

\bibitem{nuscenes2019}
H.~Caesar, V.~Bankiti, A.~H. Lang, S.~Vora, V.~E. Liong, Q.~Xu, A.~Krishnan,
  Y.~Pan, G.~Baldan, and O.~Beijbom, ``{nuScenes: A Multimodal Dataset for
  Autonomous Driving},'' \emph{arXiv:1903.11027}, 2019.

\bibitem{MOT16}
A.~Milan, L.~Leal-Taix\'{e}, I.~Reid, S.~Roth, and K.~Schindler, ``{MOT}16: {A}
  benchmark for multi-object tracking,'' \emph{arXiv:1603.00831}, 2016.

\bibitem{MOTChallenge20}
P.~Dendorfer, H.~Rezatofighi, A.~Milan, J.~Shi, D.~Cremers, I.~Reid, S.~Roth,
  K.~Schindler, and L.~Leal-Taix\'{e}, ``{MOT20: A benchmark for multi object
  tracking in crowded scenes},'' \emph{arXiv:2003.09003}, 2020.

\bibitem{openpcdet2020}
{OpenPCDet Development Team}, ``{OpenPCDet: An Open-source Toolbox for 3D
  Object Detection from Point Clouds},''
  \url{https://github.com/open-mmlab/OpenPCDet}, 2020.

\bibitem{CasTrack}
{Hai Wu}, ``{3D Multi-Object Tracker},''
  \url{https://github.com/hailanyi/3D-Multi-Object-Tracker}, 2021, {("CasTrack"
  on KITTI Leaderboard)}.

\bibitem{luiten2020trackeval}
A.~H. Jonathon~Luiten, ``{TrackEval},''
  \url{https://github.com/JonathonLuiten/TrackEval}, 2020.

\bibitem{shi2020_pvrcnn}
S.~Shi, C.~Guo, L.~Jiang, Z.~Wang, J.~Shi, X.~Wang, and H.~Li, ``{PV-RCNN:
  Point-Voxel Feature Set Abstraction for 3D Object Detection},'' in
  \emph{Conf. on Comp. Vision and Pattern Rec.}, 2020, pp. 10\,529--10\,538.

\bibitem{shi2019pointrcnn}
S.~Shi, X.~Wang, and H.~Li, ``{PointRCNN: 3D Object Proposal Generation and
  Detection from Point Cloud},'' in \emph{Conf. on Comp. Vision and Pattern
  Rec.}, 2019, pp. 770--779.

\bibitem{yan2018second}
Y.~Yan, Y.~Mao, and B.~Li, ``{SECOND: Sparsely Embedded Convolutional
  Detection},'' \emph{Sensors}, vol.~18, no.~10, p. 3337, 2018.

\bibitem{lang2019pointpillars}
A.~H. Lang, S.~Vora, H.~Caesar, L.~Zhou, J.~Yang, and O.~Beijbom,
  ``{PointPillars: Fast Encoders for Object Detection from Point Clouds},'' in
  \emph{Conf. on Comp. Vision and Pattern Rec.}, 2019, pp. 12\,697--12\,705.

\bibitem{Luiten2020}
J.~Luiten, A.~Osep, P.~Dendorfer, P.~Torr, A.~Geiger, L.~Leal-Taixe, and
  B.~Leibe, ``{HOTA: A Higher Order Metric for Evaluating Multi-Object
  Tracking},'' \emph{Int. Journal of Comp. Vision}, 2020.

\bibitem{bernardin2008}
K.~Bernardin and R.~Stiefelhagen, ``{Evaluating Multiple Object Tracking
  Performance: The CLEAR MOT Metrics},'' \emph{EURASIP Journal on Image and
  Video Processing}, pp. 1--10, 2008.

\bibitem{ravikiran2020}
M.~Ravikiran, Y.~Nonaka, and N.~Mariyasagayam, ``{A Sensitivity Analysis (and
  Practitioners’ Guide to) of DeepSORT for Low Frame Rate Video},'' in
  \emph{IEEE Int. Conf. on Big Data}, 2020, pp. 5227--5236.

\bibitem{henning2022_threat}
M.~Henning, J.~Strohbeck, M.~Buchholz, and K.~Dietmayer, ``{Identification of
  Threat Regions From a Dynamic Occupancy Grid Map for Situation-Aware
  Environment Perception},'' in \emph{IEEE Int. Conf. on Intell. Transp.
  Syst.}, 2022, pp. 805--810.

\bibitem{Zhou2022}
J.~Zhou, M.~Hirano, and Y.~Yamakawa, ``{High-Speed Rec. of Pedestrians out of
  Blind Spot with Pre-detection of Potentially Dangerous Regions},'' in
  \emph{IEEE Int. Conf. on Intell. Transp. Syst.}, 2022, pp. 945--950.

\end{thebibliography}

\end{document}